\pdfoutput=1

\documentclass[11pt]{article}

\usepackage[]{EMNLP2022}

\usepackage{times}
\usepackage{latexsym}

\usepackage{adjustbox}
\usepackage{times}
\usepackage{soul}
\usepackage{url}
\usepackage{graphicx}
\usepackage{amsmath}
\usepackage{amsthm}
\usepackage{amssymb}
\usepackage{booktabs}
\usepackage{algorithm}
\usepackage{algorithmic}
\usepackage{multirow}
\usepackage{color}
\usepackage{CJKutf8}

\usepackage[T1]{fontenc}

\usepackage[utf8]{inputenc}

\usepackage{microtype}

\usepackage{inconsolata}

\title{Contrastive Learning enhanced Author-Style Headline Generation}

\author{Hui Liu \thanks{\ \ These authors contributed equally to this work.} $^1$, Weidong Guo \footnotemark[1] \thanks{\ \ Corresponding author.} $^1$ , Yige Chen $^2$ Xiangyang Li $^1$ \\
\text{$^1$Platform and Content Group, Tencent} \\
\text{$^2$College of Computer Science and Artificial Intelligence, Wenzhou University
} \\
\texttt{$^1$\{pvopliu,weidongguo,xiangyangli\}@tencent.com} \\
\texttt{$^2$yigechen@wzu.edu.cn}
}

\begin{document}
\maketitle
\begin{abstract}
Headline generation is a task of generating an appropriate headline for a given article, which can be further used for machine-aided writing or enhancing the click-through ratio. 
Current works only use the article itself in the generation, but have not taken the writing style of headlines into consideration. 
In this paper, we propose a novel Seq2Seq model called CLH3G (\textbf{C}ontrastive \textbf{L}earning enhanced \textbf{H}istorical \textbf{H}eadlines based \textbf{H}eadline \textbf{G}eneration) which can use the historical headlines of the articles that the author wrote in the past to improve the headline generation of current articles. 
By taking historical headlines into account, we can integrate the stylistic features of the author into our model, and generate a headline not only appropriate for the article, but also consistent with the author's style. 
In order to efficiently learn the stylistic features of the author, we further introduce a contrastive learning based auxiliary task for the encoder of our model.
Besides, we propose two methods to use the learned stylisic features to guide both the pointer and the decoder during the generation. 
Experimental results show that historical headlines of the same user can improve the headline generation significantly, and both the contrastive learning module and the two style features fusion methods can further boost the performance.
\end{abstract}

\section{Introduction}

Natural Language Generation tasks have achieved great success both in research and application, such as Neural Machine Translation \cite{bahdanau2014neural}, Headline Generation \cite{jin2020hooks,ao2021pens} and so on.
In many real-life reading scenarios, an attractive headline of the article can immediately grab the readers and then lead them to view the whole article.
Thus, headline generation (HG) is becoming an important task and draw increasing attention nowadays, which aims to automatically generate the appropriate headline for a given aritcle.

Earlier research of HG \cite{dorr2003hedge,tan2017neural} mainly focus on generating a fluent and relevant headline for a given news article to alleviate the author's work in a machine-aided writing way.
More recent works \cite{zhang2018question,xu2019clickbait,jin2020hooks} intend to generate attractive headlines for articles so as to get higher click-through ratio and further directly improve the profit of the online social media platforms.
Furthermore, some works \cite{liu2020diverse,ao2021pens} try to generate keyphrase-aware and personalized headlines to meet the requirements of different application scenarios and further satisfy users' personal interests.

\begin{figure}[t]
  \centering
  \includegraphics[width=0.49\textwidth,trim=61 195 64 76,clip]{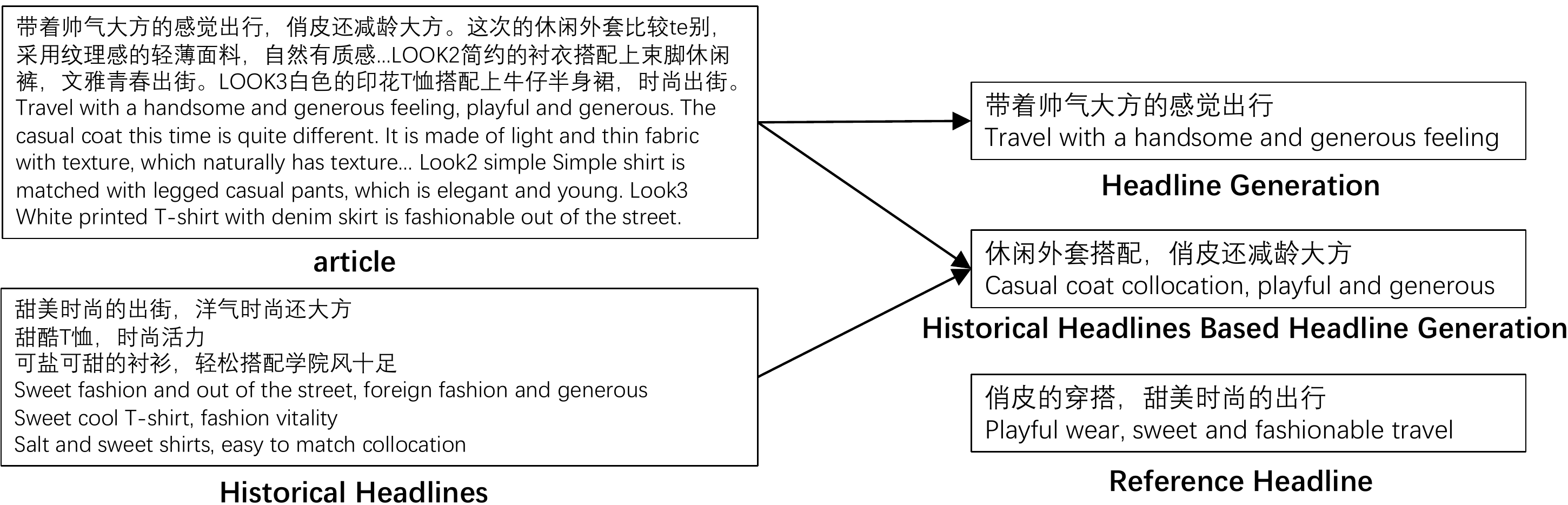}
  \caption{Comparison of general Headline Generation and Historical Headlines based Headline Generation.}
  \label{fig:example}
\end{figure}

However, most of these existing methods only use the information of the article to generate the headline but ignore the author's historical headlines.
In general, the headlines manually designed by the author are usually more suitable for the author's articles but the design style of the headline may be quite different from the writing style of the article and we can hardly obtain the author's headline style only through the content of the article.
Therefore, when we integrate existing historical headlines into the HG model to learn the headline style of the author, such as grammar and syntax, the model can generate more appropriate headlines for machine-aided writing.
For example, as shown in Figure~\ref{fig:example}, all of the historical headlines and the reference headline are composed of two clauses.
The generated headline of the historical headlines based HG model has the same syntax with the historical headlines, which makes it more likely to be accepted by the author and more attractive than the generated headline with distinctive syntax by the general HG model.

To our best knowledge, there is no corpus that contains both news articles and corresponding authorships to meet the requirement of our experiments.
Therefore, in this paper, we build a new dataset named H3G(\textbf{H}istorical \textbf{H}eadlines Based \textbf{H}eadline \textbf{G}eneration) to explore the research of historical headlines based HG.
We collect the H3G dataset from the online social media platform Tencent QQBrowser, which contains more than 380K news articles from more than 23K different authors.
The detailed introduction of the H3G dataset is discussed in the Experiment section.

Besides, we propose a novel Contrastive Learning enhanced Historical Headlines based Headline Generation (CLH3G) model to extract and learn headline styles for HG.
Inspired by the existing style transfer models \cite{lample2018multiple,dai2019style}, we represent the headline style of the historical headlines as a single vector.
Such a design can not only reduce the computation cost of the historical headlines representation, but also facilitate the integration of historical headlines information on the decoder side of the HG model.
Besides, two different methods are applied to guide the generation of the author-style headlines through the single headline vector.
The first style vector fusion method can instruct the decoder to generate author-style target headline representation, and the other controls the generated words of the pointer-generator network.
What's more, on the encoder side of Sequence-to-Sequence (Seq2Seq) model, we also use Contrastive Learning (CL) to distinguish headlines from different authors as an auxiliary task, which is consistent with and conduce to the extraction of the headline style.
%
%

%
Experimental results on automatic metrics ROUGE and BLEU and Human evaluation show that the historical headlines can greatly improve the effectiveness of HG compared with general HG models, and both of the two style vector fusion methods and Contrastive Learning based auxiliary task can also improve the performance.
We also train a Contrastive learning classifier to distinguish headlines from different authors, and find our CLH3G model can generate more author-style headlines than the general HG models and other compared models.

To this end, our main contributions are summarized as follows:
\begin{itemize}
\item We propose a new HG paradigm namely Historical Headlines based HG to generate author-style headlines, which can be used for machine-aided writing and click-through ratio enhancing.
\item We propose a novel model CLH3G, which utilizes two headline style vector fusion methods and contrastive learning to make full use of historical headlines.
\item We construct a new Historical Headlines based HG dataset namely H3G and conducted abundant experiments on it. Experimental results show that the historical headlines are beneficial to headline generation, and both the two headline style vector fusion methods and Contrastive Learning can also improve the HG models.\footnote{Our code is available at https://github.com/pvop/CLH3G}
\end{itemize}

\section{Related Work}
\textbf{Headline Generation} focus on generating a suitable or attractive headline for a given article.
We divide HG into three categories, namely general HG, style-based HG and adaptive HG.

The general HG models want to generate a fluent and suitable headline given an article.
An early work \cite{dorr2003hedge} uses linguistically-motivated heuristics to generate a matching headline.
This method is very safe, because all words in the generated headline are selected from the original article.
Then, some works \cite{see2017get,gavrilov2019self} use End-to-End neural networks to generate headlines.
These methods achieve the state-of-the-art results and are very convenient for training and inference.
Besides, \cite{tan2017neural} uses a coarse-to-fine model to generate headlines for long articles.
%
\begin{figure*}[t]
\centering
\includegraphics[width=0.8\textwidth,trim=10 80 10 40,clip]{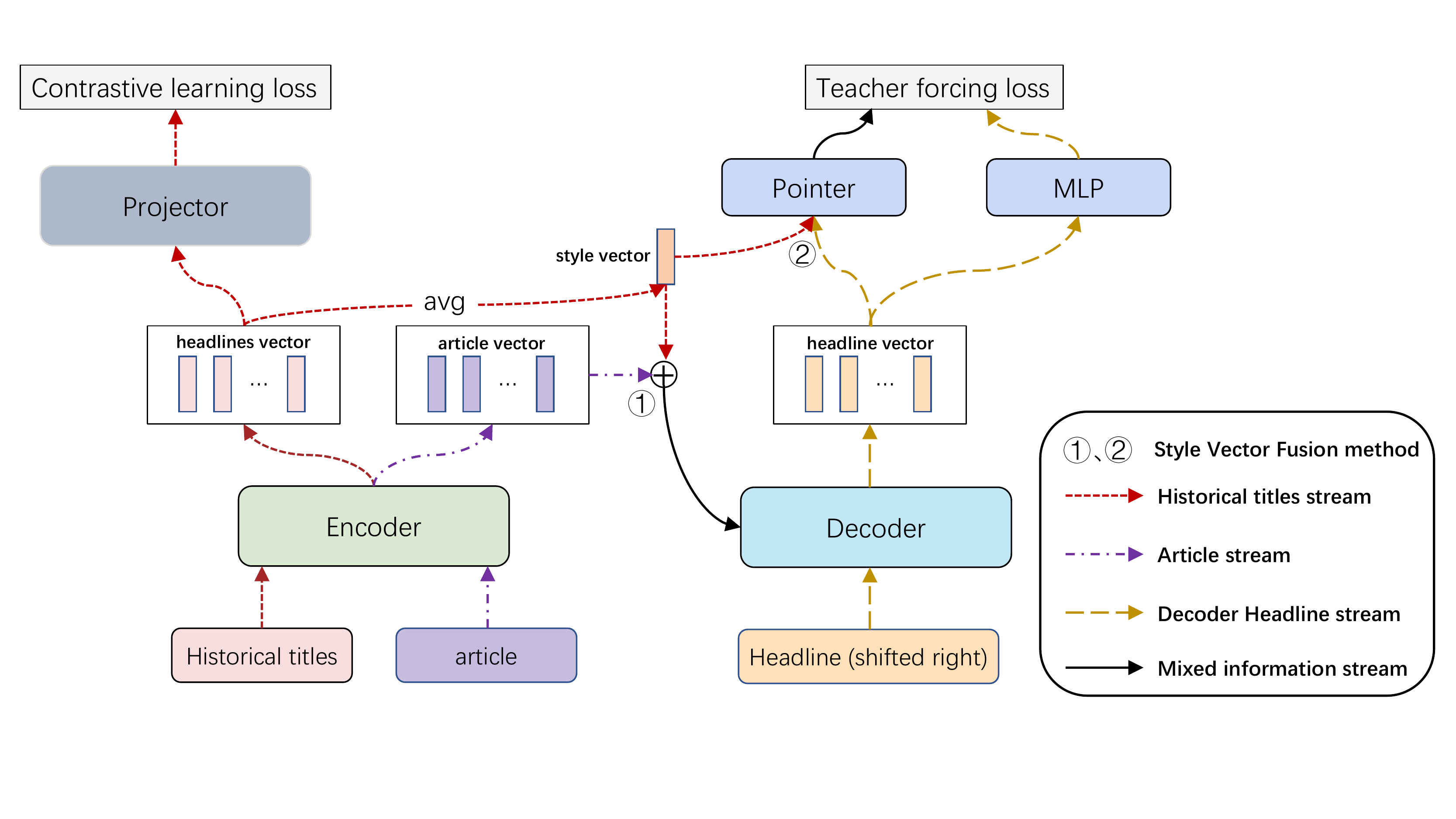} 
\caption{Our proposed Contrastive Learning enhanced Historical Headlines based Headline Generation Model.}
\label{fig:system}
\end{figure*}

The style-based headline generation models aims to generate headline with specific styles.
\cite{xu2019clickbait} uses Reinforcement Learning to generate sensational headlines to capture reader's interest.
\cite{zhang2018question} proposes dual-attention sequence-to-sequence model to generate question-style headlines, because they find question-style headlines can get much higher click-through ratio.
Besides, \cite{jin2020hooks} uses parameter sharing scheme to generate general, humorous, romantic, click-baity headlines at the same time.
%

Adaptive headline generation models want to generate different headlines for different scenarios.
\cite{liu2020diverse} proposes to generate different headlines with different keywords, which can be used to generate different headlines for different search queries in search engines.
\cite{ao2021pens} uses the user impression logs of news to generate personalized headlines for different users to satisfy their different interests.
%

\textbf{Contrastive Learning} is very popular recently for representation learning.
CL was first used for vision understanding in \cite{chen2020simple}.
Subsequently, CL is also used in Natural Language Generation, including Conditional Text Generation \cite{lee2020contrastive}, Dialogue Generation \cite{cai2020group}, Report Generation \cite{yan2021weakly} and text summarization \cite{liu2021simcls}.
In this paper, we use CL like \cite{chen2020simple}, whose framework includes a neural network encoder and a small neural network projection.

\section{Model}

Figure~\ref{fig:system} shows our proposed Contrastive Learning enhanced H3G (CLH3G) model, which is an End-to-End Seq2Seq generation model.
We will briefly introduce the entire model in Section 3.1 and discuss the encoder and the CL based auxiliary task in Section 3.2.
Finally, the decoder and two headline style vector fusion methods are presented in Section 3.3.

\subsection{Problem and Architecture}
Given an article and $k$ headlines from other articles written by the same author, our model will generate a headline which is most suitable for this article and consistent with the headline writing style of the author.
Formally, the CLH3G model uses the article $A = {[w_1^A, w_2^A,...,w_a^A]}$ of the author $X$ and some historical headlines $T = {[t_1, t_2,...,t_k]}$ of $X$ to automatically generate a new headline $H = {[w_1^H, w_2^H,...,w_h^H]}$, which is suitable for $A$ and consistent with the headline writing style of $X$.

Compared with previous HG methods, our model put more emphasis on learning the style of the input historical headlines to improve the performance.
Specifically, during encoding, we use a single vector like \cite{lample2018multiple,dai2019style} to derive the style information from the input headlines, and adopt CL to further distinguish the style among different authors.
The CL module will not bring overhead because it shares the same encoder with the original HG model.
Besides, we fuse two different methods to integrate the style information into the decoder: the first one is designed to influence the representation of the generated headline, and the other will guide the pointer module to copy author-style headline words.
In the rest of this section, we will introduce the CLH3G model in detail.

\subsection{Encoder and Contrastive Learning based Auxiliary Task}
Transformer Seq2Seq Model \cite{vaswani2017attention} has achieved remarkable success in Natural Language Generation.
Transformer consists of a self-attention multi-head encoder and a self-attention multi-head decoder.
In order to enhance the semantic representation capability of the encoder, we use the pre-trained BERT-base model \cite{devlin2018bert} to initialize the parameters of the encoder, which can generate superior article representation and headline vectors to improve the effectiveness of HG models.
%
%

As shown in Figure~\ref{fig:system}, the encoder represents the article $A$ as $H_A\in R^{a\times d}$, where $d$ is the hidden size of BERT-base.
For each headline $t_i$ in $T$, we use the encoder outputs at $[CLS]$ as its headline representation, so all historical headlines $T$ are represented as $H_T\in R^{k\times d}$.
Subsequently, we average the historical representation $H_T$ to obtain a single style vector $s_t\in R^{1\times d}$, and $H_T$ is also used to compute the CL loss.

Contrastive learning is a self-supervised method that can learn knowledge from unlabeled data.
Recently, CL has achieved great success in many fields \cite{chen2020simple,liu2021simcls}.
Same as \cite{chen2020simple}, our CL module consists of a neural network base encoder and a small neural network projection.
The CL encoder and the CLH3G encoder share parameters, so that we only need to compute the headlines representation once for both headline style vector and CL loss function to avoid additional overhead.
The projection is a two-layer fully connected feed-forward network.
Instead of explicitly constructing positive examples like most CL models, we regard the headline pairs belonging to the same author as positive samples, and the other headlines in the same batch belonging to negative samples.
The loss function of the positive pair of examples ($i$, $j$) is defined as
\begin{equation}\label{1}
  \begin{split}
  L_{i,j} = -log\frac{exp(sim(z_i,z_j)/\tau )}{\sum_{k=1}^{2N}\mathbb{I}_{[k\neq i]}exp(z_i,z_k)/\tau }
  \end{split}
\end{equation}
 where $\mathbb{I}_{[k\neq i]}$ is an indicator function evaluated to 1 iff $k\neq i$ and $\tau$ is a temperature parameter.

In the H3G dataset, all train and test samples contain at least one historical headline.
Certainly, the CLH3G model can generate the general headline only with article information, but we mainly want to explore the performance of the model when adding historical headlines.
During the training phase, the target headline is also used in CL module but not in the computation for the single style vector.
We randomly select two headlines from the input historical headlines and target headline for CL loss function.
During the inference phase, we do not use the CL module and the target headline, so that the CL module will not affect the inference speed.
\subsection{Decoder and Two Headline Style Vector Fusion Methods}
%

As shown in Figure~\ref{fig:system}, the shifted right target headline $H_{rh}$ is imputed into the decoder to generate the target headline representation matrix $D_H\in R^{h\times d}$.
During the computation of decoder, our first headline style vector fusion method is simply concatenating the article representation $H_A$ and the single style vector $s_t$ to $H_{At} \in R^{(a+1)\times d}$.
This concatenated result $H_{At}$ can guide the decoder to represent the shifted right target headline to generate a new headline with the same style of headlines in $T$.
There are many overlapping words in the headline and the corresponding article, so we use the pointer module same as \cite{see2017get} to solve the out-of-vocabulary (OOV) problem and improve the performance of generation models.
Our second headline style vector fusion method is to add the single style vector to the pointer module. 
On the one hand, we use the style vector to select words of the input article for $i-th$ generated word as:
\begin{equation}\label{1}
  \begin{split}
  \alpha (i) = \text{softmax} (w_{alpha1}^T[H_{A}:S_t] + \\ w_{alpha2}^T[d_H^i:s_t] + b_{alpha})
  \end{split}
\end{equation}
where $S_t\in R^{a\times d}$ is the result of $s_t$ repeating a times, and $w_{alpha1}$, $w_{alpha2}$ and $b_{alpha}$ are learnable parameters.
We use the headline vector $d_H^i$ to produce the vocabulary distribution of the $i-th$ generated word as:
\begin{equation}\label{1}
  \begin{split}
  P_{vocab} (i) = \text{softmax}(V^Td_H^i + b_{vocab})
  \end{split}
\end{equation}
where $V_T$ and $b_{vocab}$ are learnable parameters.
On the other hand, the generation probability $P_{gen}(i)\in[0,1]$ is computed by $H_A$, $d_H^i$ and the single style vector $s_t$ as:
\begin{equation}\label{1}
  \begin{split}
  P_{gen} (i) = \sigma (w_{gen1}^Th_{A} + w_{gen2}^Td_H^i + w_{gen3}^Ts_t+ \\ b_{gen}), \ 
  \text{where} \ h_{A}=\sum_{j=0}^{a-1}\alpha(i)_j*H_{A_j}
  \end{split}
\end{equation}
where $w_{gen1}$, $w_{gen2}$, $w_{gen3}$, $b_{gen}$ are learnable parameters. The final probability distribution of the generated word $i$ as a certain word $w$ is:
\begin{equation}\label{1}
  \begin{split}
  P_w(i) = &p_{gen}(i)P_{vocab_w}(i) + \\
  &(1-p_{gen}(i))\sum_{j:w_j=w}\alpha(i)_j
  \end{split}
\end{equation}
The final probability distribution is used to compute the teacher forcing loss, and the final loss function is:
\begin{equation}\label{1}
  \begin{split}
  Loss = L_{teacher\_forcing} + \lambda L_{contra\_learning}
  \end{split}
\end{equation}
where $\lambda$ is a hyperparameter.

The two headline style vector fusion methods take different ways to influence the final headline generation.
For a new article, there are many reasonable headlines for the article from content to syntax.
The first one is similar to informing the decoder the desired headline style in advance, allowing the decoder to have a more explicit generation direction.
Besides, headlines with different author styles have different word preferences.
The second one can guide the choice of words in the pointer network and whether to use pointer or generator.

\section{Experiments}
\subsection{Dataset}
We collect our H3G dataset from an online social media platform Tencent QQBrowser.
Some platform accounts are shared by more than one authors and publish a large number of articles every day.
So we select the accounts who published 3-60 articles within two months in 2021.
Finally, we get more than 380K different articles of more than 23K different authors, and the statistics of the H3G dataset are shown in Table 1.
\begin{table}[t]
\centering
\resizebox{1.0\columnwidth}{!}{
\begin{tabular}{cccccc}
\hline
\multirow{1}{*}{\textbf{Dataset}} & \multirow{1}{*}{\textbf{\#author}} & \multirow{1}{*}{\textbf{\#article}} & \multirow{1}{*}{\textbf{\#article per author}} & \multirow{1}{*}{\textbf{avg article length}} & \multirow{1}{*}{\textbf{avg headline length}} \\ \hline
       
\multirow{1}{*}{H3G}              & \multirow{1}{*}{23726}                  & \multirow{1}{*}{384868}              & \multirow{1}{*}{16}                                     & \multirow{1}{*}{1291}                        & \multirow{1}{*}{27}                        \\
                    
\hline
\end{tabular}
}
\caption{The statistics of H3G dataset.}
\end{table}
We randomly divide the H3G dataset into training set, validation set and test set.
The validation set and test set contain 500 and 2000 samples respectively, and the rest of the articles are used as the training set.
For these three sets, we search the historical headlines of the same author from the headlines in the training set, which avoids the answers leakage of the validation set and the test set.
In this paper, we do not consider the time when the article was published, so the historical headlines are all headlines within two months from the same author, excluding the target headline.
\subsection{Baselines}
We select two competitive models as our basic baseline models, namely general HG and merge H3G.
\begin{itemize}
\item General HG model uses transformer architecture and BERT-base to initialize the encoder parameters as our CLH3G. 
Different from our CLH3G model, the general HG model only use the original article to generate the corresponding headline. 
The general HG model is used to verify the effectiveness of historical headlines for headline generation.
\item Compared with General HG model, the merge H3G model concatenates historical headlines and the article as the input of encoder, which is a very simple method to utilize the historical headlines and can be used to verify the effectiveness of our proposed CLH3G model.
\end{itemize}

Besides, we also implement two strong baseline models, namely AddFuse HG, StackFuse HG from \cite{liu2020diverse}.
\begin{itemize}
\item The AddFuse HG model concatenates all historical headlines into a sentence as the input of the encoder to get $H_{headlines}$.
$H_{headline}$ and $H_A$ are used to compute headline-filtered article $H_{FA}$ through the multi-head self-attention sub-layer.
Finally, the target headline is generated by $H_{FA}$ instead of $H_A$.
\item Based on the AddFuse HG model, the StackFuse HG model performs a multi-head attention on $H_{FA}$ and $H_A$ one by one in each block of the decoder, so each decoder stack is composed of four sub-layers.
\end{itemize}

\subsection{Implementation and Hyperparameters}
We set the maximum article length and target headline length as 512 and 32 for all models.
The length of concatenated headlines in the AddFuse H3G model and the StackFush H3G model is 256.
The length of each historical headline in the CLH3G model is 32.
In order to be consistent with the real online applications, the number of historical headlines is random chosen from 1 to $min(K, \#(articles\ of\ the\ author)-1)$, where $K$ is a hyperparameter.
The encoder and the decoder of all transformer-based models have the same architecture hyperparameters as BERT-base.
The parameters of all models are trained by Adafactor optimizer \cite{shazeer2018adafactor}, which can save the storage and converge faster.
At the same time, we set batch size and dropout of all models to 96 and 0.1, respectively.
We train all the models 50K steps and then test on the validation set every 500 steps.
We finally report the results of the test set in the best step of the validation set.
During inference, we also use beam search with length penalty to generate more fluent headlines.
We set the beam size and length penalty of all models to 4 and 1.5, respectively.
\begin{table}[t]
\centering
\resizebox*{0.98\linewidth}{!}{
\begin{tabular}{ccccc}
\hline
\multirow{1}{*}{Model}         & \multirow{1}{*}{Rouge-1}        & \multirow{1}{*}{ROUGE-2}        & \multirow{1}{*}{ROUGE-L}       & \multirow{1}{*}{BLEU}           \\ \hline
                    
\multirow{1}{*}{General HG}    & \multirow{1}{*}{42.39}          & \multirow{1}{*}{29.29}          & \multirow{1}{*}{40.60}          & \multirow{1}{*}{22.48}          \\
                                                 
\multirow{1}{*}{Merge H3G}     & \multirow{1}{*}{42.48}          & \multirow{1}{*}{29.29}          & \multirow{1}{*}{40.42}          & \multirow{1}{*}{22.81}          \\
                                             
\multirow{1}{*}{AddFuse H3G}   & \multirow{1}{*}{43.64}          & \multirow{1}{*}{30.28}          & \multirow{1}{*}{41.59}          & \multirow{1}{*}{23.69}          \\
                                    
\multirow{1}{*}{StackFuse H3G} & \multirow{1}{*}{43.80}          & \multirow{1}{*}{30.43}          & \multirow{1}{*}{41.77}          & \multirow{1}{*}{23.76}          \\
                                     
\multirow{1}{*}{CLH3G}         & \multirow{1}{*}{\textbf{44.15}} & \multirow{1}{*}{\textbf{30.77}} & \multirow{1}{*}{\textbf{42.12}} & \multirow{1}{*}{\textbf{24.13}} \\
                               \hline
\end{tabular}
}
\caption{Rouge and BLEU scores of different Headline Generation Models.}
\end{table}

\subsection{Experimental Results on ROUGE and BLEU}
We use ROUGE \cite{lin2004rouge} and BLEU \cite{papineni2002bleu} as metrics to automatically evaluate the quality of generated headlines of all baseline models and our CLH3G model.
For all historical headlines based HG models, we set the maximum number of historical headlines $K$ to 10, and the $\lambda$ of our CLH3G model to 0.1.
As shown in Table 2, most of the H3G models are better than the general HG model, which demonstrates the effectiveness of historical headlines in headlines generation.
The merge H3G model has some similar results with the general HG model, because we only truncate the article to keep the input length of the merge H3G model as 512, and the long historical headlines causes the loss of the article information.
Besides, the two strong baseline models AddFuse HG model and StackFuse HG model achieve excellent results compared with the general HG model and the merge H3G model.
There are two main reasons for this: (1) these two models can obtain additional historical headlines information than the general HG model; (2) compared with the merge H3G model, the information of the original article will not be lost when using historical headlines.
Compared with the AddFuse H3G model, the StackFuse H3G model uses the original article representations $H_A$ incrementally and perform better results.
Finally, our CLH3G model achieves the best results for all metrics, which demonstrates our CLH3G can extract and utilize the information of historical headlines effectively compared with other baseline models.
Besides, the complexity per layer of the self-attention model is $O(n^2\cdot d)$, and our CLH3G model represents all historical headlines one by one, while the AddFuse H3G model and the StackFuse model represents all concatenated historical headlines at the same time, so our CLH3G model is more efficient than AddFuse H3G model and StackFuse model.

\subsection{Experiment results on headline style}
To study the style relationship between the generated headlines and the historical headlines, we use BERT-base and Contrastive Learning to train a classifier to distinguish headlines from different authors.
The setting of the classification model is same as the contrastive learning based auxiliary task in our CLH3G model.
The samples in the training set is a set of headlines of the same author, and we randomly select two of them as the positive samples to train the contrastive learning classifier.
%
The two headlines of the negative sample in the validation set and the test set are randomly selected from different authors.
We train the contrastive learning based classifier for 50K steps and obtain the best model in the validation set according to accuracy.
%
The contrastive learning classifier will output a score within $[-1, 1]$, and the higher the score is, the greater the possibility that the two samples belong to the same author.
We make the generated headline and all the historical headlines to build the evaluation samples one by one and report the accuracy and the average classification score.
We name the original author headline and the historical headlines as Reference, which will get the highest accuracy and average score in theory. 

The classification accuracy and the average scores are shown in Table 3.
The Reference gets the highest accuracy and average score compared with other HG and H3G models.
The general HG model obtains the worst accuracy and average score, which is consistent with its performance on ROUGE and BLEU, because it can only generate headlines aimlessly without the information of historical headlines.
We notice that the merge H3G model achieves the best accuracy and average score besides Reference.
This may be because the merge H3G model exploits the whole historical headlines and a small portion of the article to generate a new headline, and the missing information of the article makes the model relies more on historical headlines.
Compared with the general HG model, the AddFuse H3G model and the StackFuse H3G model get better results, which is also consistent with its performance on ROUGE and BLEU.
Our CLH3G model get approximate results compared with the merge H3G model, and is better than the AddFuse H3G model and the StackFuse H3G model.
The results of the classification accuracy and average score can reflect the effectiveness of using historical headlines.
Our contrastive learning module and two headline vector fusion methods are both beneficial to learn the style of historical headlines, resulting better accuracy and average score.
The best results on ROUGE and BLEU of our CLH3G model prove that our CLH3G model can utilize and fuse the article and the historical headlines effectively at the same time.

\begin{table}[t]
\centering
\resizebox*{0.98\linewidth}{!}{
\begin{tabular}{ccc}
\hline
\multirow{1}{*}{Model}         & \multirow{1}{*}{Accuracy}       & \multirow{1}{*}{Average Score}   \\ \hline
                             
\multirow{1}{*}{General HG}    & \multirow{1}{*}{86.09}          & \multirow{1}{*}{48.61}                                   \\
                      
\multirow{1}{*}{Merge H3G}     & \multirow{1}{*}{90.18}          & \multirow{1}{*}{54.76}                                     \\
                           
\multirow{1}{*}{AddFuse H3G}   & \multirow{1}{*}{89.93}          & \multirow{1}{*}{52.51}                                      \\
                            
\multirow{1}{*}{StackFuse H3G} & \multirow{1}{*}{89.41}          & \multirow{1}{*}{52.26}                                     \\
                         
\multirow{1}{*}{CLH3G}         & \multirow{1}{*}{90.13}          & \multirow{1}{*}{54.10}                      \\
                         
\multirow{1}{*}{Reference}     & \multirow{1}{*}{\textbf{92.22}} & \multirow{1}{*}{\textbf{58.21}}                       \\
             \hline
\end{tabular}
}
\caption{The contrastive learning classification results and Human Evaluation results of different Headline Generation Models.}
\end{table}

\subsection{Human Evaluation}
Besides, we also apply Human Evaluation to verify the generated headine style. 
We randomly sampled 50 news from the test set and asked three annotators to rerank the five generated headline and the reference headline, while the ranked first get 6 points, and the ranked last get 1 point.
Besides, The similar headlines will get the same ranked points, resulting the relatively high scores for all models.
We use three criteria namely fluency, relevance and attraction as \cite{jin2020hooks}.

The results is shown in the Table 4. 
Similarly with the results on Rouge, BLEU and the CL based classification, the general HG get the worst results, and the reference get the best results.
The historical headlines based models get significantly better results than the general HG model on fluency and relevance.
The historical headlines can guide the generation of target headline syntax, resulting better fluency.
Meanwhile, the better relevance is because these historical headlines based models have less factual consistency errors than the general HG model.
Finally, our CLH3G model get the best results on all three aspects except the Reference headlines.
\begin{table}[t]
\resizebox{1.0\columnwidth}{!}{
\begin{tabular}{ccccc}
\hline
\multirow{1}{*}{Model}         & \multirow{1}{*}{Fluency}      & \multirow{1}{*}{Relevance}     & {\multirow{1}{*}{Attraction}} \\ \hline
                          
\multirow{1}{*}{General HG}    & \multirow{1}{*}{4.44}         & \multirow{1}{*}{4.84}          & \multirow{1}{*}{5.14}                            \\
                         
\multirow{1}{*}{Merge H3G}     & \multirow{1}{*}{5.18}         & \multirow{1}{*}{5.14}          & \multirow{1}{*}{4.96}                            \\
                           
\multirow{1}{*}{AddFuse H3G}   & \multirow{1}{*}{4.74}         & \multirow{1}{*}{5.16}          & \multirow{1}{*}{\textbf{5.28}}                            \\
                            
\multirow{1}{*}{StackFuse H3G} & \multirow{1}{*}{4.82}         & \multirow{1}{*}{5.14}          & \multirow{1}{*}{5.26}                            \\
              
\multirow{1}{*}{CLH3G}         & \multirow{1}{*}{\textbf{5.2}} & \multirow{1}{*}{\textbf{5.18}} & \multirow{1}{*}{\textbf{5.28}}                            \\
                
\multirow{1}{*}{Reference}     & \multirow{1}{*}{\textbf{5.8}} & \multirow{1}{*}{\textbf{5.52}} & \multirow{1}{*}{\textbf{5.78}}                            \\
                                \hline
\end{tabular}
}
\caption{Human evaluation results on fluency, relevance and attraction.}
\end{table}
\begin{figure}[t]
\centering
\includegraphics[width=0.45\textwidth, trim=0 0 0 40,clip]{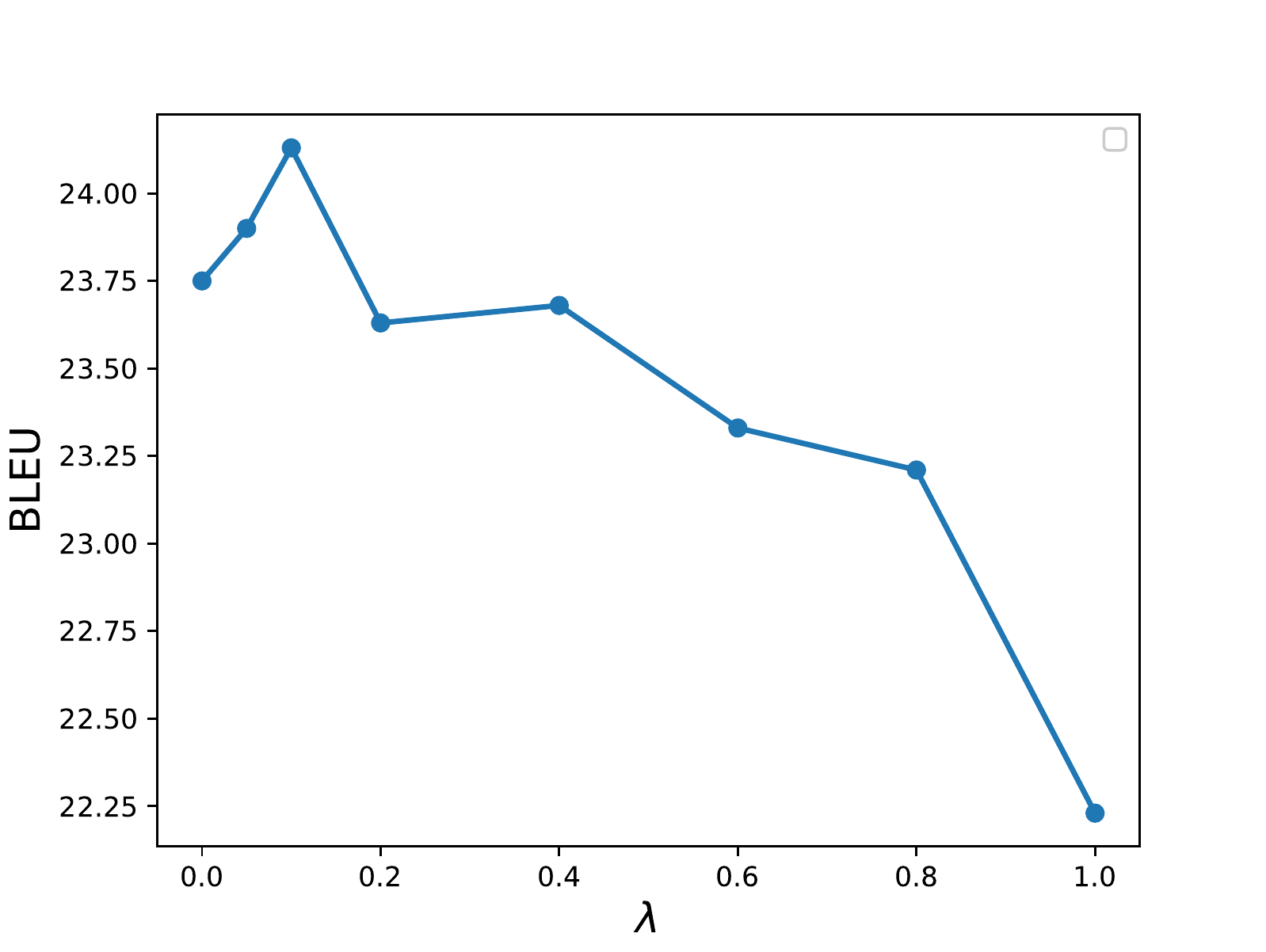} 
\caption{Results of CLH3G with different values of Contrastive Learning coefficient $\lambda$.}
\label{fig:result-coefficient}
\end{figure}
\subsection{Experimental Results with different values of hyperparameter $\lambda$}
In order to study the influence of contrastive learning module, we train our CLH3G model with different contrastive learning coefficient $\lambda$.
We report the BLEU results of the experiments in Figure~\ref{fig:result-coefficient}.
The different values of $\lambda$ have a great impact on the final results of our CLH3G model, and a clear conclusion can be drawn from the results.
When $\lambda$ is smaller than 0.1, the larger of $\lambda$, the better of the performance.
And when $\lambda$ is bigger than 0.1, the smaller of $\lambda$, the better of the performance.
The best result is achieved when $\lambda$ is 0.1.
We will analyze the reasons of this experiment results.
Firstly, when $\lambda$ is very small, contrastive learning module has little positive impact on the whole model, so that the results are getting better.
Then, with the increasing of $\lambda$, contrastive learning module has too much impact on the whole model, which disturbs the training of headline generation, so the results are getting worse.
\begin{CJK}{UTF8}{gbsn}
\begin{table*}[t]
\resizebox{\textwidth}{!}{
\begin{tabular}{p{480pt}}
\hline
\begin{tabular}[c]{@{}p{480pt}@{}}\textbf{Historical Headlines}:\\ 1.  Digital currency has been accepted by the world, and the Chinese market is about to explode! (数字货币已被世界公认，中国市场即将爆发！)\\ 2. The central bank takes strong action, and RMB is about to rise! (人民币崛起 央行强势出手！)\\ 3. Digital RMB develops from "point" to "surface", and has entered thousands of families! (数字人民币由“点”到“面”，走进千家万户！)\end{tabular} \\ \hline
\begin{tabular}[c]{@{}p{480pt}@{}}\textbf{Reference headline}:\\ The central bank is launching digital RMB, indicating the full outbreak of digital currency! (央行数字人民币落地，迎来数字货币全面爆发！)\end{tabular}   \\ \hline
\begin{tabular}[c]{@{}p{480pt}@{}}\textbf{Generated headline by general HG model}:\\ Daofu Chen, Financial Research Institute of the State Council: China should give better play to the cross-border payment system based on blockchain (国金融研所陈道富：中国宜更好发挥基于区块链等的跨境支付体系的作用)\end{tabular}            \\ \hline
\begin{tabular}[c]{@{}p{480pt}@{}}\textbf{Generated headline by CLH3G model}:\\ The central bank is launching digital RMB, striving to become the first player in the field! (央行数字人民币即将落地，争做第一个吃螃蟹的人！)\end{tabular}  \\ \hline
\end{tabular}
}
\caption{An Example of generated headlines through general HG model and CLH3G model.}
\end{table*}
\end{CJK}

\begin{table}[t]
\resizebox{1.0\columnwidth}{!}{
\begin{tabular}{lcccc}
\hline
\multirow{1}{*}{Model}                       & \multirow{1}{*}{Rouge-1}        & \multirow{1}{*}{Rouge-2}        & \multirow{1}{*}{Rouge-L}        & \multirow{1}{*}{BLEU}           \\ \hline
                                      
\multirow{1}{*}{General HG}                  & \multirow{1}{*}{42.39}          & \multirow{1}{*}{29.29}          & \multirow{1}{*}{40.60}          & \multirow{1}{*}{22.48}          \\
                                                             
\multirow{1}{*}{+ Concat Style Vector}       & \multirow{1}{*}{43.63}          & \multirow{1}{*}{30.31}          & \multirow{1}{*}{41.79}          & \multirow{1}{*}{23.90}          \\
                                       
\multirow{1}{*}{+ Concat Style Vector + CL}  & \multirow{1}{*}{43.66}          & \multirow{1}{*}{30.39}          & \multirow{1}{*}{41.45}          & \multirow{1}{*}{23.74}          \\
                  
\multirow{1}{*}{+ Pointer Style Vector}      & \multirow{1}{*}{43.09}          & \multirow{1}{*}{29.93}          & \multirow{1}{*}{41.25}          & \multirow{1}{*}{23.20}
          \\
\multirow{1}{*}{+ Pointer Style Vector + CL} & \multirow{1}{*}{43.55}          & \multirow{1}{*}{30.24}          & \multirow{1}{*}{41.82}          & \multirow{1}{*}{23.35}          \\
                                           
\multirow{1}{*}{+ Two fusion Methods}        & \multirow{1}{*}{43.37}          & \multirow{1}{*}{30.32}          & \multirow{1}{*}{41.58}          & \multirow{1}{*}{23.75}          \\
                                       
\multirow{1}{*}{+ Two fusion Methods + CL}   & \multirow{1}{*}{\textbf{44.15}} & \multirow{1}{*}{\textbf{30.77}} & \multirow{1}{*}{\textbf{42.12}} & \multirow{1}{*}{\textbf{24.13}} \\ \hline
\end{tabular}
}
\caption{Incremental Experiment of CLH3G model.}
\end{table}

\subsection{Incremental Experiments}
To further demonstrate the effectiveness of contractive learning module and the two headline vector fusion methods in our CLH3G model, we conduct incremental experiments and report the results in Table 4.
%
%
As shown in Table 6, the concat and pointer headlines style vector fusion methods both can improve the performance of Headline Generation, because they use additional historical headlines.
In addition, the concat fusion method can get better results compared with the pointer fusion method, which proves that informing the decoder the desired headline style in advance is more effective than guiding the choice of words in pointer network.
It may also be due to that there are few overlapping words between different headlines of the same author, and their headline patterns and style are consistent instead.
We also add contrastive learning based auxiliary task to the concat fusion method and the pointer fusion method, respectively.
The performance of the concat fusion method using CL based auxiliary task is slightly improved in ROUGE-1 and ROUGE-2, while the performance in ROUGE-L and BLEU is reduced, which shows that CL has little effect on the concat fusion method.
The pointer fusion method with the CL based auxiliary task greatly improves the effectiveness of all metrics, which proves that the pointer is more dependent on the headline style.
Furthermore, when we use the two fusion methods at the same time, the results is somewhere in between using a single method, because the relatively worse headline vector will mislead the choice of words in the pointer.
For our CLH3G model, the better headline vector leads to the best results, which demonstrates our contrastive learning module can extract better headline style vector for H3G models.

\subsection{Case Study}
We display an example of generated headlines by general HG model and CLH3G model in Table 5.
Both the historical headlines and the generated headline by CLH3G model are exclamatory sentences.
Besides, the generated headline by CLH3G is more informative and attractive than the generated headline by general HG model. 


\section{Conclusion}
In this paper, we discuss the effectiveness of Historical Headlines for Headline Generation, and aim to generate headlines not only appropriate for the given news articles, but consistent with the author's style.
We build a large Historical Headlines based Headline Generation dataset, and propose a novel model CLH3G to integrate the historical headlines effectively, which contains a contrastive learning based auxiliary task and two headline style vector fusion methods.
Experimental results show the effectiveness of historical headlines for headline generation and the exceptional performance of both the CL based auxiliary task and the two headline style vector fusion methods of our CLH3G model.

\section*{Limitations}
This paper introduces a new headline generation task, which use historical headlines to generate article headlines, and also proposes a novel model called CLH3G for this task.
CLH3G uses two headline style vector fusion methods to make full use of historical headlines.
However, those two style vector fusion methods is difficult to applied into pretrained Sequence to Sequence model including T5, Mass \cite{song2019mass, raffel2019exploring} directly, because those two methods will change the whole architecture of pretrained model, resulting slightly worse results compared with original pretrained model.
As result, the integration of CLH3G and pretrained Sequence to Sequence models requires abundant H3G data to achieve comparable results. 
\section*{Acknowledgements}
We would like to thank the anonymous reviewers for their constructive comments.
\bibliographystyle{acl_natbib}
\bibliography{custom}
\end{document}